\documentclass{article}

\usepackage{arxiv}

\usepackage[utf8]{inputenc} 
\usepackage[normalem]{ulem}

\usepackage[T1]{fontenc}    
\usepackage{hyperref}       
\usepackage{url}            
\usepackage{booktabs}       
\usepackage{amsfonts}       
\usepackage{nicefrac}       
\usepackage{microtype}      
\usepackage{lipsum}
\usepackage{graphicx}
\usepackage{multirow} 
\usepackage{wrapfig,float}
\usepackage{subcaption}
\graphicspath{ {./images/} }
\usepackage{amsmath}

\usepackage{placeins}

\usepackage[ruled,vlined,linesnumbered]{algorithm2e} 
\SetKwInput{KwIn}{Input}\SetKwInput{KwOut}{Output}   

\usepackage{xcolor} 
\definecolor{darkpurple}{rgb}{0.5, 0.2, 0.8}
\definecolor{darkyellow}{rgb}{0.5, 0.5, 0.0}
\definecolor{darkblue}{rgb}{0.0, 0.0, 0.8}

\title{Constrained Tabular Diffusion for Finance}

\author{
  Michael Cardei\textsuperscript{*} \\
  University of Virginia\\
  \texttt{ntr2rm@virginia.edu} \\
  \And
  Jose M Munoz\textsuperscript{*} \\
  Massachusetts Institute of Technology\\
  \texttt{josemm@mit.edu} \\
  \And
  Oscar Barrera \\
  Harvard University\\
  \texttt{oscarbarrera@g.harvard.edu} \\
  \And
  Shreyas K Chandrahas \\
  Visa Inc\\
  \texttt{shrchand@visa.com} \\
  \And
  Partha Saha \\
  Visa Inc\\
  \texttt{pasaha@visa.com} \\
}

\begin{document}
\maketitle
\begingroup
\renewcommand\thefootnote{\fnsymbol{footnote}}
\footnotetext[1]{Equal and main contribution.}
\endgroup
\pagestyle{plain}
\begin{abstract}
Generative models in finance face the dual challenge of producing realistic data while satisfying strict regulatory and economic objectives, a requirement that standard tabular diffusion models cannot provide. To address this difficulty, we introduce \textbf{Constrained Tabular Diffusion for Finance} (CTDF), a novel integration of sampling-time feasibility operations with mixed-type tabular diffusion in financial applications. By incorporating a training-free feasibility operator into the reverse‑diffusion sampling loop, CTDF enforces hard constraints for applications such as simulation, legal compliance, and extrapolation. Extensive experiments on large-scale financial datasets demonstrate \emph{zero constraint violations} and improvement in scarce data utility. CTDF establishes a robust method for generating trustworthy and compliant synthetic data, opening new avenues for rigorous generative modeling and analysis in the financial domain.

\end{abstract}

\section{Introduction}\label{sec:intro}

High-quality synthetic data is essential for the modern financial ecosystem, from global payment networks to digital banking and commerce platforms. Within this domain, synthetic data has become a vital tool for developing robust fraud detection systems, calibrating credit risk models, and enabling privacy-compliant data sharing~\cite{lu2025machinelearningsyntheticdata}. However, the utility of this data is contingent upon two core requirements: it must not only capture the complex statistical patterns of real-world distributions but also adhere to the complex landscape of business logic, economic principles, and regulatory guardrails governing financial operations. Failure to enforce these constraints invalidates model outputs, undermines strategic decision-making, and exposes an institution to significant compliance and operational risk. Conversely, the ability to generate data that is both statistically faithful and structurally sound allows for discovery, enabling more nuanced simulations of market dynamics and deeper insights into the drivers of business performance.

Early work in financial data synthesis relied on deep generative models, such as Generative Adversarial Networks (GANs) \cite{eckerli2021generativeadversarialnetworksfinance, pu2016variational} and Variational Autoencoders (VAEs) \cite{potluru2024syntheticdataapplicationsfinance}. Recent advances have shifted toward diffusion probabilistic models, whose likelihood‑based training and iterative denoising yield better mode coverage and sample fidelity. While these approaches provide mixed-type high-fidelity data generation \cite{zhang2024tabsyn, shi2025tabdiff, kotelnikov2023tabddpm, sattarov2023findiffdiffusionmodelsfinancial}, their stochastic sampling remains unconstrained, unable to enforce adherence to strict constraints.

Existing approaches to controlling generative output often fall short in the financial domain. Model conditioning, such as classifier-free guidance, can only influence outputs without strictly enforcing them~\cite{ho2022classifierfreediffusionguidance}. An alternative, post hoc correction, involves generating a full batch of samples and then filtering out invalid ones or projecting them onto the feasible set in a single final step. This approach is not only computationally inefficient, as it requires sampling significantly more, but can compromise the data’s statistical integrity, leading to a significant divergence from the learned distribution that the generative model was trained to replicate \cite{robert1999monte}.

\begin{figure}
    \centering
    \includegraphics[width=0.7\linewidth]{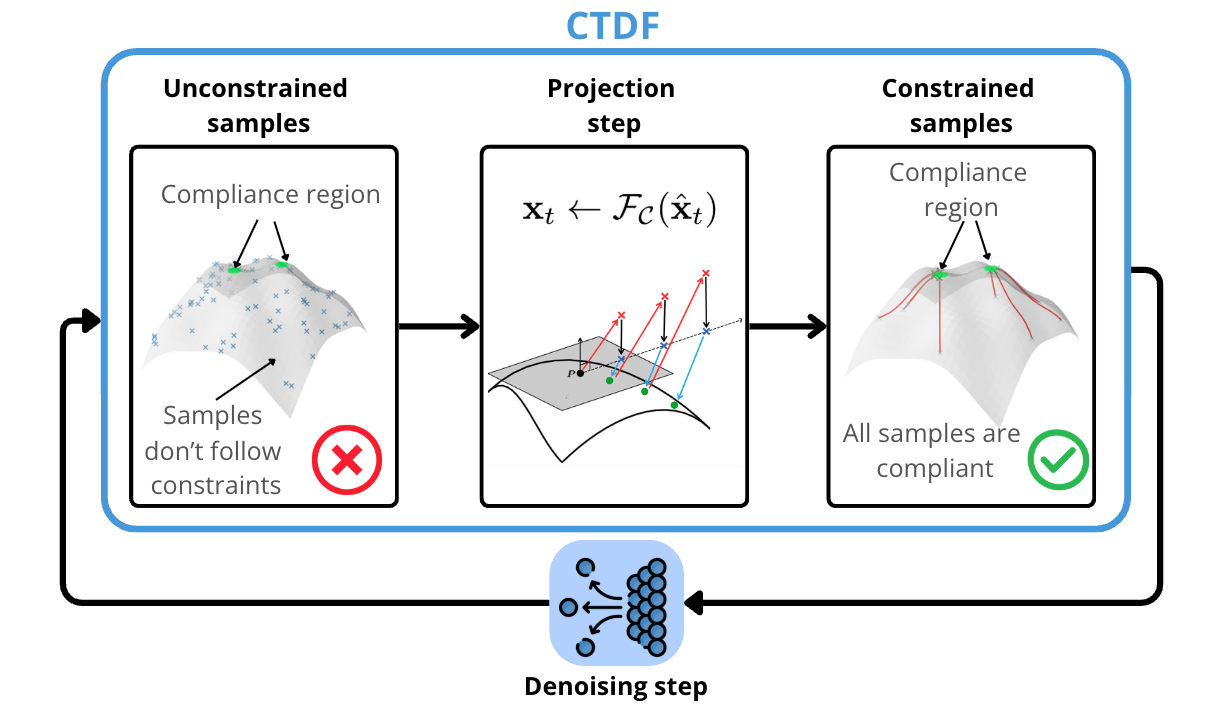}
    \caption{CTDF Enforces Constraints via the feasibility operation. At each step, unconstrained samples are mapped onto the compliance region, ensuring all generated outputs satisfy domain-specific constraints.}
    \label{fig:enter-label}
\end{figure}

To address these problems, we introduce \textbf{Constrained Tabular Diffusion for Finance (CTDF)}. After every denoising update in the reverse diffusion chain, CTDF enforces samples to lie inside the feasible region. This acts as a training-free feasibility operator inside the sampling loop, actively steering the denoising trajectory, ensuring that every sample remains within the feasible region defined by hard constraints. Notably, as the mapping occurs throughout the trajectory, rather than a post‑hoc fix, the synthetic samples stay aligned with the base model’s learned joint structure. Thus, this approach preserves sample fidelity while enforcing compliance.

This study provides the following contributions:

(I) We introduce CTDF, the first training-free, constraint enforcing framework that integrates feasibility mapping into mixed‑type tabular diffusion for financial applications.

(II) We design a {hybrid mapping operator} that seamlessly handles the mixed data types of financial records by combining Euclidean mapping for numerical features with KL-divergence-based mapping for categorical features.

(III) We experimentally demonstrate on two large-scale financial datasets that CTDF achieves \textbf{zero violations} on hard constraints while preserving high data fidelity.

\section{Related Work}

Diffusion models have recently emerged as a dominant approach for tabular data synthesis. TabDDPM introduced separate kernels for numerical and categorical features \cite{kotelnikov2023tabddpm}, while subsequent models such as STaSy proposed self-paced curricula to enhance sample fidelity \cite{kim2023stasy}. To improve efficiency, TabSyn performed diffusion in a compressed latent space learned via a VAE encoder \cite{zhang2024tabsyn}. TabDiff further addressed feature heterogeneity by employing feature-wise learnable noise schedules, enabling a single continuous-time model to adapt to the distributional characteristics of each column \cite{shi2025tabdiff}. In the financial domain, state-of-the-art diffusion models have demonstrated growing promise. However, their stochastic sampling procedures provide no guarantees of adherence to the stringent business rules and regulatory requirements inherent to financial data. To mitigate this, FinDiff incorporated domain-specific architectures and financial pretraining to improve synthesis quality on structured financial datasets \cite{sattarov2023findiffdiffusionmodelsfinancial}. However, it still lacks mechanisms for enforcing strict constraints and cannot guarantee regulatory compliance during sampling.

To ensure outputs follow specific attributes, recent work has focused on integrating guidance directly into the diffusion process. Diffusion-based guidance methods steer samples toward desired attributes by modifying the score. Classifier guidance augments the score with the gradient of a pretrained classifier to bias generation toward a target class \cite{dhariwal2021diffusionmodelsbeatgans}, while classifier-free guidance interpolates between conditional and unconditional scores removing the need for an external classifier \cite{ho2022classifierfreediffusionguidance}. Such guidance improves alignment with conditioning variables but does \emph{not} provide hard, per-sample feasibility guarantees. Other work handles this with the integration of constraint optimization within the reverse diffusion process by recasting sampling as a constrained optimization problem. Projected Diffusion Models (PDM) introduced this idea for continuous data, using a projection operator after each denoising step to keep samples within a feasible set \cite{christopher2024pdm}. This projection-based approach was generalized across data types with the Neurosymbolic Diffusion framework of NSD \cite{christopher2025nsd}. We build on these works with, to our knowledge, the first study to implement and evaluate per-step hard constraint enforcement in mixed-type \emph{tabular} diffusion with a focus on financial datasets and compliance-driven constraints.

\section{Preliminaries}

\paragraph{Diffusion Models.} Generative diffusion models \cite{ho2020denoisingdiffusionprobabilisticmodels,sohldickstein2015deepunsupervisedlearningusing, song2021scorebasedgenerativemodelingstochastic} have revolutionized data synthesis, generating high-fidelity, state-of-the-art image and video samples \cite{liu2024sorareviewbackgroundtechnology, rombach2022highresolutionimagesynthesislatent}. Diffusion models consist of a \emph{forward} Markov chain that gradually corrupts a clean sample $\mathbf{x}_0\sim p_{\text{data}}$ to approach the noisy distribution $\mathbf{x}_T\!\sim\!\mathcal{N}( \mathbf{0},\mathbf{I})$, followed by a learned \emph{reverse} chain that reconstructs $\mathbf{x}_0$ from $\mathbf{x}_T$.
Formally, the forward process $\{\mathbf{x}_t\}_{t=0}^T$ is determined by the kernel $q(\mathbf{x}_t\mid \mathbf{x}_{t-1}) = \mathcal{N}\!\bigl(\sqrt{1-\beta_t}\,\mathbf{x}_{t-1},\; \beta_t\mathbf{I}\bigr)$, where the variance schedule $\{\beta_t\}$ increases monotonically. The reverse process is guided by a neural network $s_\theta(\mathbf{x}_t, t)$ trained to predict the {score}, $\nabla_{\mathbf{x}_t}\log p_t(\mathbf{x}_t)$, where $p_t$ is the marginal data distribution at time $t$. This score function defines a deterministic generative process via the {probability flow ODE}. To generate a sample, one solves this ODE backwards in time from $t=T$ to $t=0$, typically with a numerical solver. A representative update step is:
$$
\mathbf{x}_{t-\Delta t} = \mathbf{x}_t + g(t) s_\theta(\mathbf{x}_t, t)\Delta t
$$
where $g(t)$ is a function of the noise schedule. This procedure deterministically transforms a noise vector $\mathbf{x}_T$ into a data sample $\mathbf{x}_0$ that approximates the true data distribution $p_{\text{data}}$.

\paragraph{Discrete Diffusion Models.} Although diffusion was initially developed for continuous data, recent works extend this concept from continuous vectors to discrete token sequences \cite{austin2023structureddenoisingdiffusionmodels, sahoo2024simpleeffectivemaskeddiffusion}. Each token is a one-hot vector $\mathbf{x}\!\in\!\{0,1\}^V$ over a vocabulary of size $V$, where a sample is a length–$L$ sequence $\mathbf{x}_0=(\mathbf{x}^{(1)}_0,\dots,\mathbf{x}^{(L)}_0)$.
The forward Markov chain replaces tokens with noise at a time-dependent rate $\beta_t$ given by: $q(\mathbf{x}_t\mid\mathbf{x}_{t-1})
=(1-\beta_t)\,\mathbf{x}_{t-1}+\beta_t\,\nu$.
Here, $\nu$ is either a uniform distribution \cite{schiff2025simpleguidancemechanismsdiscrete} or a dedicated \texttt{[MASK]} symbol \cite{sahoo2024simpleeffectivemaskeddiffusion}. The reverse process is modeled as \[
x_{t - \Delta} =
\begin{cases}
\text{Cat}\left(x_{t - \Delta};\, x_t\right), & \text{if } x_t \neq \nu \\
\text{Cat}\left(x_{t - \Delta};\,
\frac{\beta(t - \Delta)\, \nu + \left(\beta(t) - \beta(t - \Delta)\right)\, s_\theta(x_t, t)}{\beta(t)}
\right), & \text{if } x_t = \nu
\end{cases}
\]
where the learned denoiser, $s_\theta(x_t,t)$ approximates the sample $x_0$ and $x_t$ is the probability distribution over $V$ for each token in the sequence. 

\paragraph{Mixed-Type Tabular Diffusion Models.} Tabular data combines numerical $\mathbf{x}^{(\text{num})} \in \mathbb{R}^N$ and categorical $\mathbf{x}^{(\text{cat})} \in \prod_{k=1}^C \Delta_{K_k}$ features. Purely continuous diffusion fails to capture categorical structure, and purely discrete diffusion cannot represent real-valued geometry. 
Early tabular diffusion approaches tackled this by running separate processes for each feature type~\cite{kotelnikov2023tabddpm, lee2023codi}, or by first embedding every row into a continuous latent space via a VAE and then diffusing there~\cite{zhang2024tabsyn, zheng2023diffusionmodelsmissingvalue}, bypassing the need to handle categorical noise kernels directly. More recent work (\textsc{TabDiff}) \cite{shi2025tabdiff} models an entire table row $\mathbf{x}_t$ as a trajectory of a continuous-time diffusion process and assigns to each column~$j$ an independent, learnable noise schedule $\alpha_j(t)\in[0,1]$. 
A shared Transformer encoder, conditioned on the partially corrupted row $\mathbf{x}_t$ and a time embedding $\tau(t)$, produces a time-aware representation $h_t = f_{\theta}(\mathbf{x}_t, \tau(t))$. Two specialized heads then predict (i) Gaussian score estimates for numerical features and (ii) categorical logits for discrete features. During generation, a reverse step denoises the numerical features while categorical tokens are resampled from the predicted logits, enabling a single model to reconstruct mixed-type tables jointly. 

\section{Methods}
\label{sec:methods}

To generate high-fidelity tabular data that adheres to a set of predefined constraints, CTDF extends a pre-trained, unconstrained mixed-type diffusion model, which has learned the underlying data distribution. For this implementation, we use TabDiff \cite{shi2025tabdiff} as the base diffusion model architecture; however, CTDF can be extended to other tabular diffusion architectures. CTDF introduces a feasibility operation-based framework that enforces constraints during the generation process at each sampling step to ensure the final output lies within a feasible, constraint-compliant region, without requiring any retraining or fine-tuning of the base model. This approach relies on diffusion’s inherent parallel denoising, modeling all columns simultaneously, updating the current sample jointly at each step.

\paragraph{Unconstrained Reverse Process} Let $\mathbf{x} = (\mathbf{x}^{\text{num}}, \mathbf{x}^{\text{cat}})$ be a data sample, representing a financial record with $N$ numerical features $\mathbf{x}^{\text{num}} \in \mathbb{R}^N$ and $C$ categorical features $\mathbf{x}^{\text{cat}} \in \prod_{k=1}^{C} \{0,1\}^{K_k + 1}$, where $K_k$ is the number of categories for the $k$-th feature and the extra category is the special [MASK] token also represented as $\mathbf{m}$.

In order to sample  $\mathbf{x}^{\mathrm{num}}_T \sim \mathcal{N}(0, I_N )$ and $\mathbf{x}^{\mathrm{cat}}_T = \mathbf{m}$ we use the base sampling algorithm described in \cite{shi2025tabdiff}. Specifically, after sampling from the prior, a forward diffusion step to \( t^+ = t + \gamma_t \Delta t \), where \( \gamma_t \in (0, 1) \), the step-size fraction, is used to allow for self-correction by randomly re-perturbing the numerical and categorical features. Following this, we apply the reverse update, which, for numerical values, computes the probability flow ODE direction and takes one step in the reverse diffusion chain:
\[
d\mathbf{x}^{\mathrm{num}}
\;=\;
\frac{\mathbf{x}^{\mathrm{num}}_{t^+}
      - \mu^{\mathrm{num}}_{\theta}(\mathbf{x}_{t^+},t^+)}
     {\sigma_{\mathrm{num}}(t^+)},
\qquad
\mathbf{x}^{\mathrm{num}}_{t-1}
= \mathbf{x}^{\mathrm{num}}_{t^+}
  + \bigl(\sigma_{\mathrm{num}}(t-1)-\sigma_{\mathrm{num}}(t^+)\bigr)\,d\mathbf{x}^{\mathrm{num}}.
\]
where $\mu^{\mathrm{num}}_{\theta}(x,t)\in\mathbb{R}^N$ is the model’s numerical head and $\sigma_{\mathrm{num}}(t)$ is the noise schedule.  
For categorical values we sample from the masked posterior:
\[
x^{\mathrm{cat}}_{t-1}
\sim
p_{\theta}\!\bigl(x^{\mathrm{cat}}_{t-1}\mid x^{\mathrm{cat}}_{t^{+}},\,\mu^{\mathrm{cat}}_{\theta}(x_{t^{+}},t^{+})\bigr).
\]
where unmasked tokens are copied forward and masked tokens are resampled from the model’s categorical distribution.


\paragraph{Sampling-time Feasibility Operation for Constrained Generation.}

Financial applications commonly require imposing a set of hard constraints on the data, defining a feasible set $\mathcal{C} = \{\mathbf{x} \mid h_i(\mathbf{x}) = 0,\ i=1,\dots,m;\; g_j(\mathbf{x}) \le 0,\ j=1,\dots,r\}.$
A naive approach to guarantee these constraints is to perform post-hoc rejection sampling, but this is inefficient and can severely distort the learned data distribution. Instead, we extend the core idea from PDM \cite{christopher2024pdm} by integrating the constraints directly into the generative process. Hence, CTDF generates samples that approximate the conditional distribution
\(p_\theta(\mathbf{x}\mid \mathbf{x}\in\mathcal{C})\) while satisfying all constraints during sampling.

At each step $t$ of the reverse SDE, after computing the unconstrained update $\mathbf{x}_{t-\Delta t}$,
we apply a \textbf{feasibility operator}, $\mathcal{F}_{\mathcal{C}}$, which maps the unconstrained update onto the feasible set $\mathcal{C}$. This operator is integrated directly into the sampling loop: $$ \mathbf{x}_{t-1} \leftarrow \mathcal{F}_{\mathcal{C}}(\tilde{\mathbf{x}}_{t-1}) $$ By repeatedly applying this operator, we map the entire sampling trajectory to remain within the feasible region, guiding the generation towards a valid and high-fidelity output.

\begin{wrapfigure}[20]{r}{0.5\textwidth}
    \vspace{-2em} 
    \begin{algorithm}[H]
        \caption{Constrained Tabular Diffusion (CTDF)}
        \label{alg:ctdf_general}
        \small
        \DontPrintSemicolon
        \LinesNotNumbered
        
        \KwIn{Pre-trained model $\mu_\theta=(\mu_\theta^\mathrm{num}, \mu_\theta^\mathrm{cat})$, schedules $\{\sigma_t\}, \{\alpha_t\}, \{\gamma_t\}$, unified feasible set $\mathcal{C}$, steps $T$}
        \KwOut{Constraint-compliant sample $\mathbf{x}_0 \in \mathcal{C}$}
        
        Sample $\mathbf{x}_T^{\mathrm{num}} \sim \mathcal{N}(\mathbf{0}, I_N)$; $\mathbf{x}_T^{\mathrm{cat}} \leftarrow \mathbf{m}$\;
        
        \For{$t=T, \dots, 1$}{
            $t^+ \leftarrow t + \gamma_t \Delta t$\;
            Sample $\boldsymbol{\epsilon} \sim \mathcal{N}(\mathbf{0}, \mathbf{I})$\;
            $\mathbf{x}_{t^+}^{\mathrm{num}} \leftarrow \mathbf{x}_t^{\mathrm{num}} + \sqrt{\sigma(t^+)^2 - \sigma(t)^2} \boldsymbol{\epsilon}$\;
            Sample $\mathbf{x}_{t^+}^{\mathrm{cat}} \sim q(\cdot | \mathbf{x}_t^{\mathrm{cat}}, 1 - \alpha_{t^+}/\alpha_t)$\;
            $\mathbf{x}_{t^+} \leftarrow (\mathbf{x}_{t^+}^{\mathrm{num}}, \mathbf{x}_{t^+}^{\mathrm{cat}})$\;
            
            $d\mathbf{x}^{\mathrm{num}} \leftarrow (\mathbf{x}_{t^+}^{\mathrm{num}} - \mu_\theta^{\mathrm{num}}(\mathbf{x}_{t^+}, t^+)) / \sigma(t^+)$\;
            $\tilde{\mathbf{x}}_{t-1}^{\mathrm{num}} \leftarrow \mathbf{x}_{t^+}^{\mathrm{num}} + (\sigma(t-1) - \sigma(t^+)) d\mathbf{x}^{\mathrm{num}}$\;
            $\hat{\mathbf{p}}_{t-1}^{\mathrm{cat}} \leftarrow \text{softmax}(\mu_\theta^{\mathrm{cat}}(\mathbf{x}_{t^+}, t^+))$\;
            
            \tcp{Unified feasibility operation}
            $\tilde{\mathbf{x}}_{t-1} \leftarrow (\tilde{\mathbf{x}}_{t-1}^{\mathrm{num}}, \hat{\mathbf{p}}_{t-1}^{\mathrm{cat}})$\;
            $(\mathbf{x}_{t-1}^{\mathrm{num}}, \mathbf{p}_{t-1}^{\mathrm{cat}}) \leftarrow \mathcal{F}_{\mathcal{C}}(\tilde{\mathbf{x}}_{t-1})$\;

            $\mathbf{x}_{t-1}^{\mathrm{cat}} \sim \mathbf{p}_{t-1}^{\mathrm{cat}}$\;
        }
        
        \Return{$(\mathbf{x}_0^{\mathrm{num}},\; \mathbf{x}_0^{\mathrm{cat}})$}
        \label{alg:only}
    \end{algorithm}
    \vspace{-1.5em}
\end{wrapfigure}

\paragraph{Feasibility Operation for Mixed-Type Financial Data.}
As financial data frequently contains both numerical and categorical features, the choice of the distance metric $D$ is critical and must be tailored to the mixed-data-type nature. The distance metric is a sum of modality-specific distances: $$D(\mathbf{x}, \hat{\mathbf{x}}) = D_{\text{num}}(\mathbf{x}^{\text{num}}, \hat{\mathbf{x}}^{\text{num}}) + D_{\text{cat}}(\mathbf{x}^{\text{cat}}, \hat{\mathbf{x}}^{\text{cat}}).$$

Numerical constraints in finance are typically affine, defining a convex polytope $\mathcal{C}^{\text{num}} = \{\mathbf{x} \in \mathbb{R}^N \mid \mathbf{A}\mathbf{x} \leq \mathbf{b}\}$. The natural distance metric for real-valued vectors is the squared Euclidean distance, $D_{\text{num}}(\mathbf{x}, \hat{\mathbf{x}}) = \|\mathbf{x} - \hat{\mathbf{x}}\|^2$. The mapping is thus:
\begin{equation*}
    \mathcal{F}_{\mathcal{C}^{\text{num}}}(\hat{\mathbf{x}}^{\text{num}}) = \operatorname*{argmin}_{\mathbf{x} \in \mathcal{C}^{\text{num}}} \|\mathbf{x} - \hat{\mathbf{x}}^{\text{num}}\|^2.
\end{equation*}

For categorical features, the diffusion model outputs a vector of logits for each column, which gets interpreted as an unnormalized log-probability distribution over the possible categories. Let $\hat{\mathbf{p}}_k = \text{softmax}(\hat{\mathbf{z}}_k)$ be the predicted probability vector for the $k$-th categorical feature.

The appropriate distance metric for probability distributions is the Kullback-Leibler (KL) divergence. The mapping finds a new distribution $\mathbf{q}_k$ supported only on the valid categories $\Omega_k$ that is closest to the model's prediction $\hat{\mathbf{p}}_k$:
\begin{equation*}
    \mathcal{F}_{\mathcal{C}^{\text{cat}}_k}(\hat{\mathbf{p}}_k) = \operatorname*{argmin}_{\mathbf{q}_k \in \Delta_{K_k}, \text{supp}(\mathbf{q}_k) \subseteq \Omega_k} \text{KL}(\mathbf{q}_k \| \hat{\mathbf{p}}_k).
\end{equation*}
The solution to this KL operation has a simple and intuitive closed form: set the probabilities of all invalid categories to zero and re-normalize the probabilities of the valid ones. This preserves the relative likelihoods assigned by the model within the valid subset. 

\paragraph{Functional and Symbolic Constraints.}
Beyond simple decoupled rules, many crucial business regulations involve coupled constraints where the valid range of one feature depends on the values of others. Our framework handles this broad class of functional constraints, which can be generalized as $y \le f(z)$, where $y$ is a target feature and $z$ is a set of one or more input features. The function $f$ is highly flexible; for instance, it can encode symbolic \texttt{if-then} logic, such as a financial rule where $\texttt{price} \ge 1000$ if $\texttt{property\_type} = \text{'Single Family'}$. Furthermore, $f$ can be a pre-trained model, such as a neural network, allowing for highly complex, non-linear constraints as used in Subsection \ref{subsec:house}. To implement these capabilities, we employ a targeted, sequential mapping. At each denoising step $t$, the feasibility operation uses the current values of the input features, $z_t$, to compute the constraint boundary $b_t = f(z_t)$. A feasibility mapping is then applied \emph{only} to the target column $y_t$ to ensure it satisfies $y_t \le b_t$, leaving the input columns $z_t$ untouched by this specific operation. This method allows CTDF to enforce complex, interdependent rules without requiring a full, computationally expensive joint projection.

\paragraph{The CTDF Sampling Algorithm.}
The CTDF sampling procedure, detailed in Algorithm \ref{alg:only}, integrates our feasibility operation directly into the reverse generative process of a base model like TabDiff. The core of our approach is an iterative loop: at each reverse timestep $t$, the base model proposes an unconstrained update via the score and probabilities. Our feasibility operation, $\mathcal{F}$, then projects this update onto the valid sets $\mathcal{C}^{\mathrm{num}}$ and $\mathcal{C}^{\mathrm{cat}}$ before the state for that step is finalized. This ensures the final output $\mathbf{x}_0$ is constraint-compliant.

\paragraph{Fidelity Metrics. }\label{sec:fidelities}

We evaluate generated data fidelity using four metrics. Marginal distribution quality is measured by average Shape Similarity; the average of Kolmogorov-Smirnov (KS) for continuous columns, and Total Variation Distance (TVD) for categorical columns. Distributional Distance consists of the average 1D-Wasserstein distance for continuous columns and Jensen-Shannon (JS) divergence for categorical columns. Bivariate dependencies are assessed via the Trend Score, which compares the pairwise Pearson's column correlations between the real and synthetic datasets for continuous columns and TVD for categorical columns. Finally, overall similarity is measured by the Detection AUC, the ROC-AUC score of a CatBoost classifier~\cite{prokhorenkova2018catboost}, which is trained to distinguish real from synthetic data, where 0.5 indicates perfect indistinguishability. The baseline benchmarks for all models were performed using their default parameters and using the same 90/10 train-test split, where the fidelity metrics were computed using the test split. All the experiments were performed using NVIDIA A30 GPUs.

\FloatBarrier
\section{Experimental Results}

\begin{table}[h!]
\centering
\renewcommand{\arraystretch}{1.2}
\resizebox{0.8\linewidth}{!}{%
\begin{tabular}{@{}lcccc@{}}
\toprule
\multirow{2}{*}{\textbf{Model}} & \multicolumn{2}{c}{\textbf{Marginal Fidelity}} & \textbf{Correlations} & \textbf{ML Similarity} \\
\cmidrule(lr){2-3} \cmidrule(lr){4-4} \cmidrule(lr){5-5}
& Shape Sim. (Avg) $\uparrow$ & Dist. Dist. (Avg) $\downarrow$ & Trend Score $\uparrow$ & Detection AUC $\to 0.5$ \\
\midrule
CTGAN~\cite{xu2019modeling}                   & 0.86483 & 0.11454 & 0.79725 & 0.99999 \\
TVAE~\cite{ishfaq2023tvae}                    & 0.84638 & 0.09050 & 0.83098 & 0.99985 \\
FinDiff~\cite{sattarov2023findiffdiffusionmodelsfinancial} & 0.94472 & 0.02916 & 0.87369 & 0.98654 \\
Tabsyn~\cite{zhang2024tabsyn} & 0.97646 & 0.02027 & 0.94595 & 0.79479 \\
\textbf{CTDF (TabDiff~\cite{shi2025tabdiff})} & \textbf{0.98787} & \textbf{0.01116} & \textbf{0.95376} & \textbf{0.67860} \\
\bottomrule
\end{tabular}%
}

\caption{Fidelity Benchmarks on the Housing Market Dataset.
Our unconstrained base model (TabDiff) is compared against standard generative models. For metric definitions see \autoref{sec:fidelities}.}
\label{tab:fidelity_benchmark}
\end{table}

\subsection{Housing market}\label{subsec:house}

To evaluate CTDF’s ability to generate realistic, constraint-compliant tabular data, we use the Airbnb Open Data from major U.S. cities, a large-scale real estate dataset with 450k property listings~\cite{seth_us_2020}. It includes a rich mix of 9 numerical attributes (e.g., price, latitude/longitude, reviews) and 5 categorical features (e.g., room type, neighborhood). This domain is especially well-suited for our setting: listings must satisfy a growing number of local housing regulations, platform-level business rules, and investment constraints, making it a natural application domain for constraint-aware generative modeling.

Before assessing constraint satisfaction, we first validate the fidelity of our underlying unconstrained diffusion model against several established benchmarks for tabular data synthesis. Table~\ref{tab:fidelity_benchmark} summarizes the performance across three key dimensions: marginal distribution fidelity, pairwise column correlations, and machine learning utility. The results demonstrate TabDiff's superior performance in all metrics, capturing distributional shapes, pairwise trends, and overall statistical patterns. This provides a high-quality, state-of-the-art starting point for our constrained generation framework.


\begin{figure*}[t]                
  \vspace{0pt}
  \centering                     
  \begin{minipage}[t]{0.57\linewidth}   
    \centering
    \includegraphics[width=\linewidth]{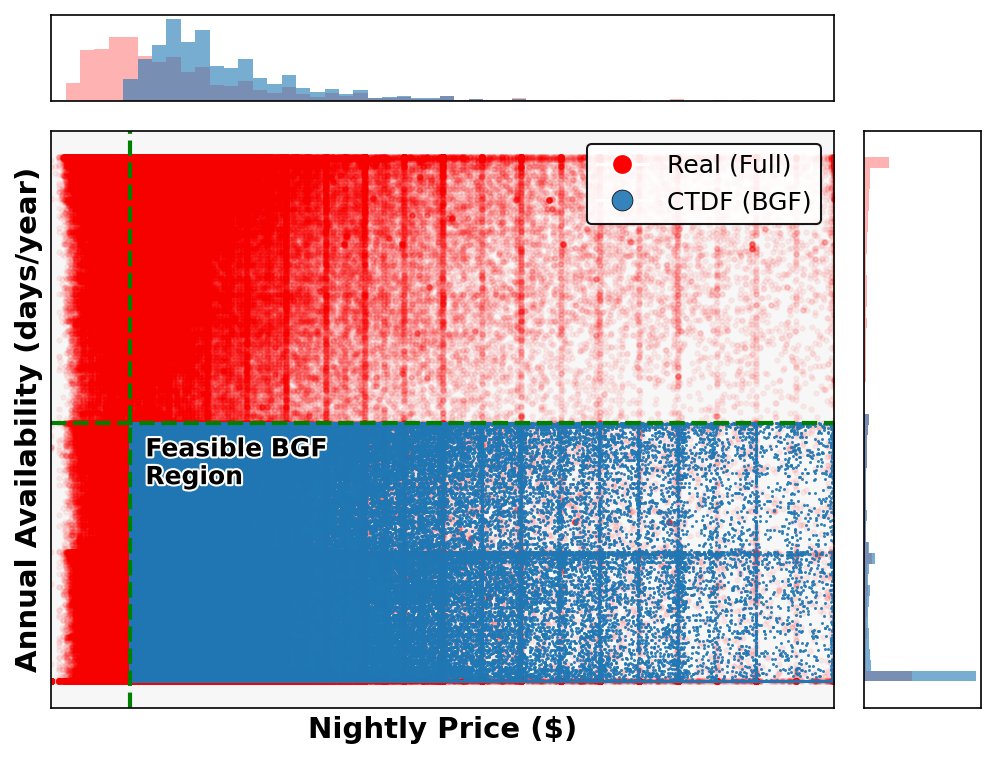}
  \end{minipage}\hfill
  \begin{minipage}[t]{0.41\linewidth}
   \vspace{-180pt}
    \centering
    \footnotesize                 
    \renewcommand{\arraystretch}{1.2}

    \begin{tabular}{@{}lrr@{}}
      \toprule
      Rule & Real (\%) & CTDF (\%)\\
      \midrule
      Listings $<2$          & 43.7 & 0.0 \\
      Price $\ge 100$        & 35.2 & 0.0 \\
      Min nights $\le 30$    & 6.5  & 0.0 \\
      Reviews/mo $\ge 1.5$   & 70.2 & 0.0 \\
      Availability $\le 180$ & 43.7 & 0.0 \\
      \textbf{All rules}     & 99.9 & \textbf{0.0} \\
      \bottomrule
    \end{tabular}


    \begin{tabular}{@{}lrrr@{}}
      \toprule
      Statistic & Real $\mu\!\pm\!\sigma$ & CTDF $\mu\!\pm\!\sigma$ \\
      \midrule
      Nightly price (\$) & 207$\pm$109 & 229$\pm$120  \\
      Availability       &  88$\pm$56 &  42$\pm$56  \\
      Reviews/mo         & 4.1$\pm$3.0& 1.9$\pm$0.7 \\
      Listings/host      & 1.6$\pm$1.0& 1.1$\pm$0.3 \\
      \bottomrule
    \end{tabular}
  \end{minipage}

  \caption{ CTDF vs.\ real Airbnb listings.\label{fig:airbnb_ctdf_vs_real}
           \emph{Left:} Scatter plot comparing real NYC listings (red) to CTDF synthetic listings for the Balanced Growth Fund (blue). Green dashed lines mark regulatory thresholds (\$100 nightly minimum, 180-day annual availability). CTDF samples fall entirely in the compliant lower-right quadrant.  
           \emph{Right:} Constraint-violation rates (top) and descriptive-statistic match (bottom).}
  \label{fig:airbnb_ctdf_composite}
\end{figure*}
\paragraph{Compliant Portfolio Simulation}
To showcase CTDF’s scenario-analysis capabilities, we construct the \textit{Balanced Growth Fund (BGF)}, a synthetic portfolio of high-end, legally compliant, entire-home rentals in New York City. For this scenario, we fix the categorical columns to \texttt{room\_type = Entire home/apt}, \texttt{city = New York City}, and impose several hard constraints. To ensure adherence to New York's housing regulations (Local Law 18 \cite{nyc_registration_law}), we enforce primary-residency rules by requiring \texttt{calculated\_host\_listings\_count < 2} and an annual \texttt{availability\_365} $\le$ 180. To reflect an actively managed, non-corporate listing, \texttt{reviews\_per\_month} is constrained between 1.5 and 15. Finally, to align with the fund's investment strategy targeting the upper-tier market, we set a minimum \texttt{price} $\ge$ \$100 and \texttt{minimum\_nights} $\le$ 30. 

Consequently, CTDF enables the simulation of portfolios that are simultaneously realistic, compliant, and strategically aligned, unlocking unprecedented precision for risk modeling and scenario planning in finance. The resulting output of CTDF is shown in Fig.~\ref{fig:airbnb_ctdf_vs_real}, where a comparison between the real housing market and the synthetic sample illustrates the compliance of two vital constraints. This experiment demonstrates how CTDF enables precise simulation of legally compliant and strategically targeted real estate portfolios, empowering financial institutions to model regulatory exposure, test investment hypotheses, and design products in alignment with evolving market rules.

\paragraph{Equitable Pricing Auditing and Algorithmic Gentrification Correction Simulation}

\begin{figure}[htbp]          
  \centering
  \includegraphics[width=1\linewidth]{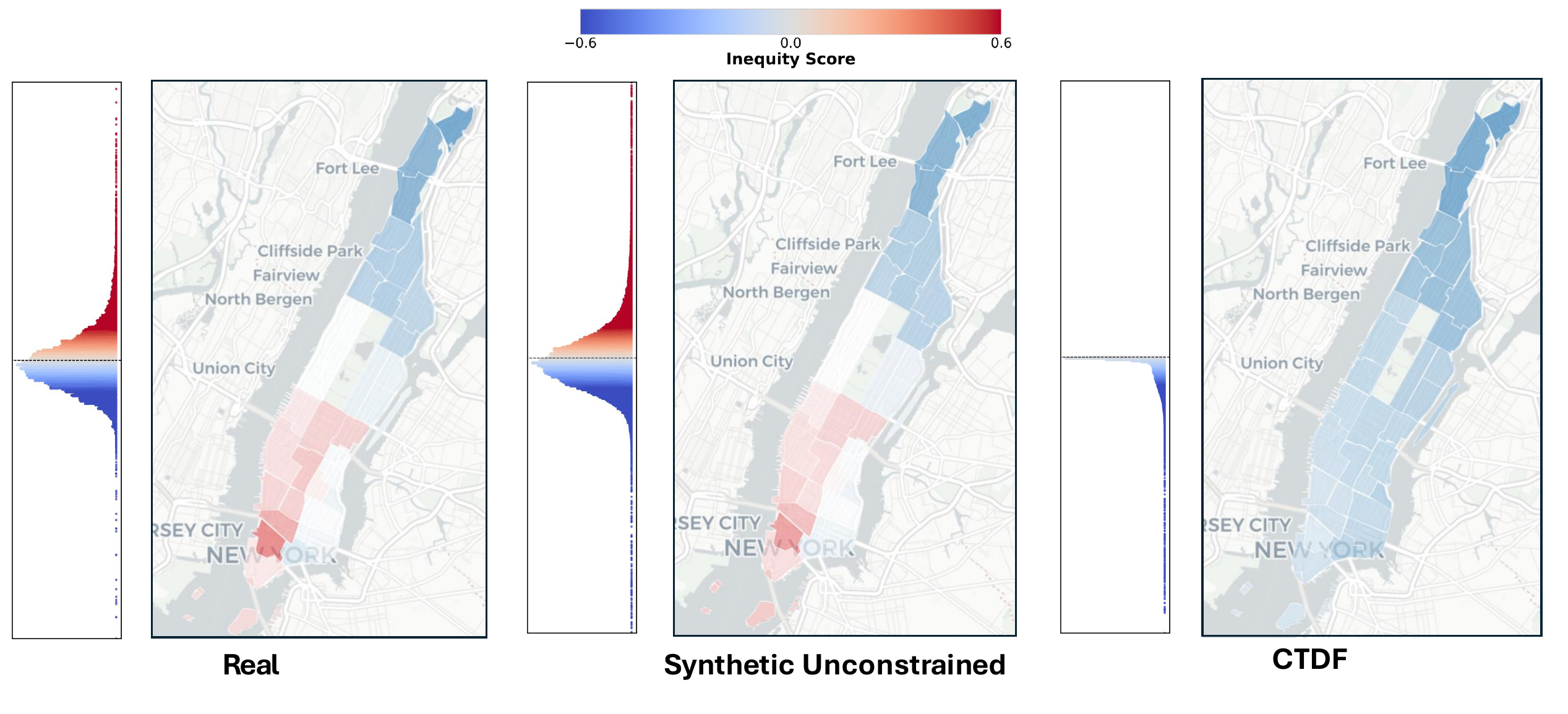}
  \caption{CTDF eliminates location-based price inequity in the NYC housing market.
The inequity score (red = overpriced, blue = underpriced) reflects the difference between a listing's actual price and a location-blind "fair" price.
\emph{Left:} The real market shows significant price inflation in certain neighborhoods.
\emph{Center:} An unconstrained generative model learns and reproduces this geographical bias.
\emph{Right:} By constraining generated prices to match fair prices, CTDF creates a synthetic dataset purged of this bias, demonstrating its utility for algorithmic fairness and auditing.
}
  \label{fig:inequity}
\end{figure}

\begin{wraptable}[10]{r}{0.48\textwidth}
  \centering
  \resizebox{1\linewidth}{!}{%
    \begin{tabular}{@{}lrrr@{}}
      \toprule
      Metric & Real & Unconstrained & CTDF \\
      \midrule
      Shape $\uparrow$                & --    & 0.988 & 0.983 \\
      Dist. Dist $\downarrow$           & --    & 0.011 & 0.022 \\
      Trend $\uparrow$                 & --    & 0.948 & 0.946 \\
      Constraint Viol. (\%) & 42.64 & 42.34   & \textbf{0.0} \\
      \bottomrule
    \end{tabular}%
  }
  \caption{Constraint violation and fidelity results for unconstrained generation and CTDF.}
  \label{tab:inequality}
\end{wraptable}

Housing price models can absorb and amplify location effects tied to historical demographic patterns, leading to systematic over- or under-pricing across neighborhoods, referred to as algorithmic gentrification. In this work, we use CTDF to (i) measure these disparities and (ii) \emph{simulate} corrective policies inside a controlled generative setting relevant to financial risk, revenue, and fairness. We train a regression network $f_{\theta}$ on \emph{intrinsic} listing attributes (amenities, size, host / review features), deliberately excluding \texttt{latitude}, \texttt{longitude}, \texttt{neighbourhood\_group}, and \texttt{neighbourhood}. Its prediction $\hat{p}^{\mathrm{IFP}} = f_{\theta}(x^{\text{intr}})$ estimates a location‑neutral (intrinsic) price. The inequity score is $e = p - \hat{p}^{\mathrm{IFP}}$, with $e>0$ interpreted as an over‑pricing premium and $e<0$ as discounted. Aggregating $e$ over neighborhoods exposes systematic disparities.  \\
An unconstrained TabDiff model is trained on the NYC subset over the full joint of intrinsic attributes, neighborhood indicators, and observed price $p$. To simulate the removal of premiums, we enforce the hard constraint $e \le 0$. During reverse diffusion, each denoising step applies a per‑step mapping on the price column, leaving all other attributes untouched. Samples already satisfying $e\le 0$ pass through unchanged; only over‑priced listings are mapped to their intrinsic surrogate. For this application, both $f_{\theta}$ and the constraint are restricted to listings with \texttt{neighbourhood\_group} = \textsc{Manhattan}; other NYC borough listings are excluded from fitting the surrogate and from constrained sampling. \\

Figure~\ref{fig:inequity} (Real, Synthetic Unconstrained, CTDF) shows that the unconstrained model faithfully reproduces the spatial over‑pricing pattern (positive tail in $e$ and red neighborhoods), indicating the disparity is learnable. Under CTDF, the positive tail is cleanly truncated while neighborhood structure and variability remain. Table~\ref{tab:inequality} confirms that constraint violations drop to $0\%$ while marginal, distributional, and dependency fidelity metrics stay near the unconstrained baseline (a small divergence increase reflects removal of the premium tail). This yields a plausible counterfactual “fair‑price’’ synthetic market stripped of location‑driven markups. This simulation demonstrates CTDF's ability to offer a transparent mechanism to audit and then simulate alternative pricing regimes via systematic constraint enforcement.

\subsection{Loan Analysis}

\begin{table}[h!]
\centering
\renewcommand{\arraystretch}{1.2}
\resizebox{0.8\linewidth}{!}{%
\begin{tabular}{@{}lcccc@{}}
\toprule
\multirow{2}{*}{\textbf{Model}} & \multicolumn{2}{c}{\textbf{Marginal Fidelity}} & \textbf{Correlations} & \textbf{ML Similarity} \\
\cmidrule(lr){2-3} \cmidrule(lr){4-4} \cmidrule(lr){5-5}
& Shape Sim. (Avg) $\uparrow$ & Dist. Dist. (Avg) $\downarrow$ & Trend Score $\uparrow$ & Detection AUC $\to 0.5$ \\
\midrule
CTGAN~\cite{xu2019modeling}                    & 0.84188 & 0.10975 & 0.80008 & 0.99997 \\
TVAE~\cite{ishfaq2023tvae}                    & 0.79720 & 0.00867 & 0.87228 & 0.99845 \\
FinDiff~\cite{sattarov2023findiffdiffusionmodelsfinancial}                 & 0.96711 & 0.00637 & 0.94790 & 0.93186 \\
Tabsyn~\cite{zhang2024tabsyn}                 & 0.99402 & 0.01338 & 0.96397 & 0.86901 \\
\textbf{CTDF (TabDiff~\cite{shi2025tabdiff})} & \textbf{0.99345} & \textbf{0.00002} & \textbf{0.96519} & \textbf{0.70553} \\
\bottomrule
\end{tabular}
}
\caption{
    Quantitative Fidelity Benchmarks on the Lending Club Dataset.
    CTDF's base model (TabDiff) is compared against standard generative baselines.  Arrows indicate the preferred direction for model quality.\label{tab:lending_club_fidelity}
}
\end{table}

We now turn to the distinct domain of consumer credit to test CTDF in a core financial domain. We use a comprehensive Lending Club dataset, which contains historical data for all peer-to-peer loans issued between 2007 and 2020~\cite{ethon0426_lendingclub_2007_2020q1}. This large-scale dataset, with nearly 3 million records, provides a granular view of borrowers' financial health, loan characteristics, and credit history. After preprocessing, our working dataset features 41 columns, representing a complex mixed-type environment. This includes 22 numerical features capturing both continuous and discrete values, such as loan amount and interest rates, alongside 19 categorical attributes that detail loan status and purpose.

\begin{wrapfigure}[30]{r}{0.5\textwidth} 
  \vspace{-0.5\baselineskip}
  \centering
  \footnotesize

  \setlength{\tabcolsep}{7pt}        
  \renewcommand{\arraystretch}{1.12} 
  \begin{tabular}{lccc}              
    \toprule
    Metric                & Real  & Unconstrained & CTDF \\
    \midrule
    Shape $\uparrow$                 & --    & 0.99344 & 0.94231 \\
    Dist.\ Dist.\ $\downarrow$       & --    & 0.01651 & 0.11876 \\
    Trend $\uparrow$                 & --    & 0.95996 & 0.92126 \\
    Viol.\ (\%)                      & 90.51 & 90.22   & \textbf{0.0} \\
    \bottomrule
  \end{tabular}

  \vspace{0.55\baselineskip}

  \includegraphics[width=\linewidth]{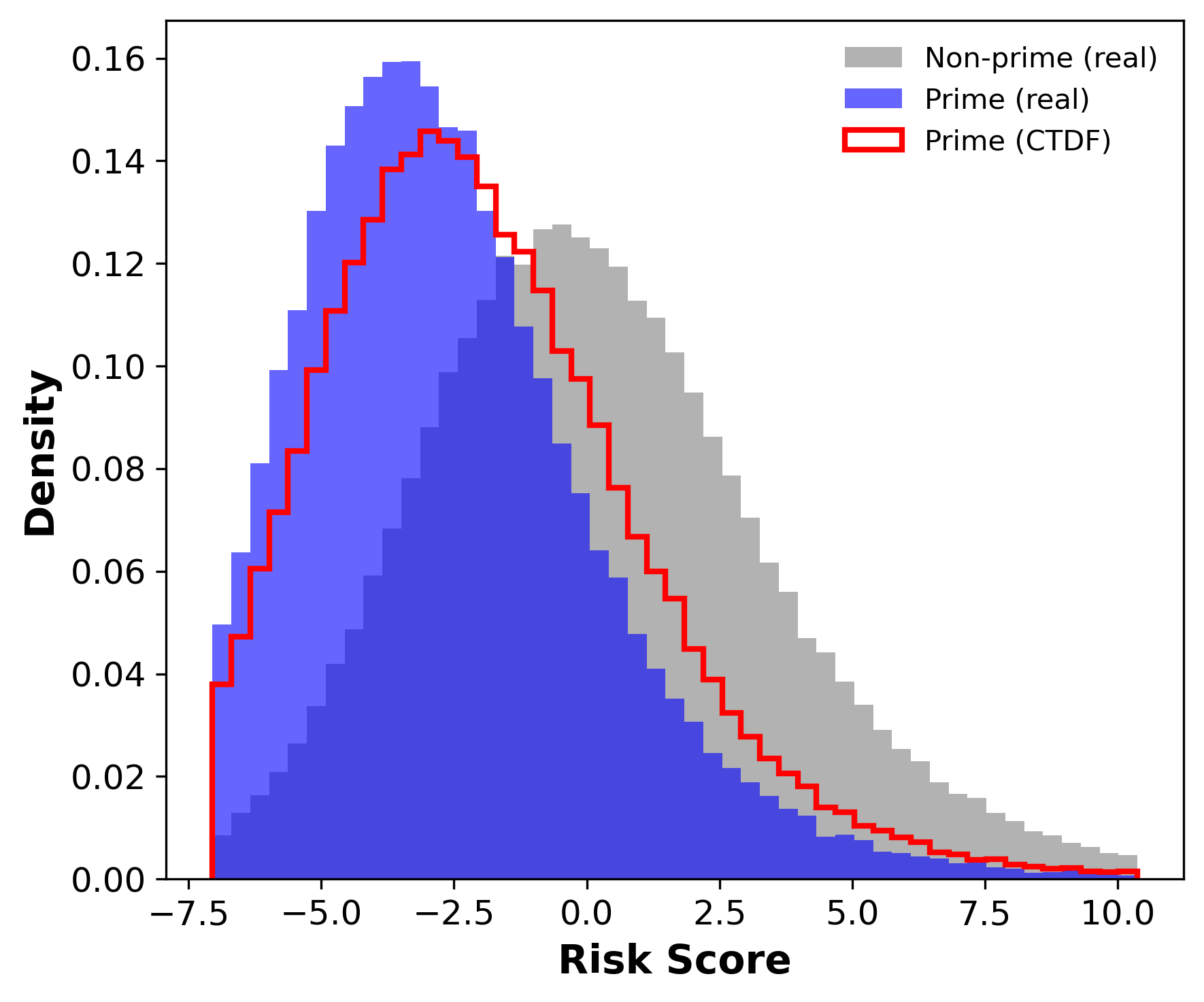}

  \caption{\textbf{Top:} Quantitative metrics (shape, distance, trend fidelity, and constraint violations).
           \textbf{Bottom:} Distribution of the composite Credit Risk Score. Lower values indicate safer borrowers.
           Non-prime loans (grey) cluster above 0; prime loans (blue) lie below; CTDF’s synthetic prime loans (red outline) track the real prime profile.}
  \label{fig:risk-score-dist}
\end{wrapfigure}

The fidelity benchmarks presented in~\autoref{tab:lending_club_fidelity} confirm that the underlying TabDiff model again establishes a state-of-the-art performance.

\paragraph{Simulating a Compliant Prime Lending Portfolio}
In consumer lending, lenders must prove that every product assignment is anchored in legitimate, measurable credit risk factors and never in prohibited attributes under laws such as the Equal Credit Opportunity Act (ECOA)~\cite{usdoj2025equalcredit}. To illustrate how a generative model can help satisfy a specific, policy-driven risk appetite, we tasked CTDF with synthesizing a portfolio drawn exclusively from the highest credit-quality segment of the market, a slice so small in the real data that traditional sampling is impractical for stress testing, capital planning, or training specialized risk models.

CTDF enforces four hard constraints on every synthetic loan: (1) \texttt{FICO\_RANGE\_LOW} $\geq$ 720 to guarantee prime-tier credit; (2) \texttt{DTI} $\leq$ 36 to cap leverage; (3) \texttt{VERIFICATION\_STATUS} $\in$ \{Verified, Source Verified\} to reduce fraud risk; and (4) \texttt{SUB\_GRADE} restricted to the A and B buckets. Only ~ 10\% of the original dataset meets all four rules simultaneously, yet CTDF generates 300k samples with zero violations, demonstrating that the model can learn and obey complex, multi-column business logic inside a narrowly defined risk band.

To gauge overall creditworthiness beyond the imposed boundaries, we construct a simple composite credit risk score by standardizing each feature, then summing the {risk-increasing} factors: interest rate, credit inquiries in the past six months, revolving-credit utilization, number of active bank-card lines, total revolving accounts, number of trade lines opened in the past 12 months, presence of a debt-settlement flag, and shorter time since the last delinquency, while subtracting the {risk-reducing} factors: annual income, percentage of accounts that have never been delinquent, and length of employment. Lower scores mark safer profiles. We then plot the distribution of this composite score for three sets: the broad non-prime population, the observed prime slice that meets all four hard constraints in the real data, and the synthetic prime loans produced by CTDF. Figure~\ref{fig:risk-score-dist} displays these distributions, showing how the synthetic prime curve closely tracks the real prime histogram while remaining well-separated from the non-prime mass.

\paragraph{AI-Driven Credit Risk for Small-Business Loans}
\label{sec:sb_ai_experiment}

Modern credit risk models rely on large, balanced datasets to train powerful machine learning scoring functions.  However, when a lender seeks to automate risk decisions for the small-business segment, the severe class imbalance (\(<1\%\) of loans) and regulatory caps on \textsc{DTI}, \(\textsc{int\_rate}\), and \(\textsc{revol\_util}\) create two AI challenges: (1) \emph{data scarcity} in the high-rate, high-risk tail, and (2) \emph{distributional bias} when unconstrained sampling produces implausible loans that corrupt the training signal. CTDF solves both by generating a large, \emph{compliant} synthetic SB portfolio, enabling AI models to learn robust patterns exactly where real data is weakest.

We enforce per-step feasibility operation on every CTDF sample $
\textsc{DTI}\le0.43,\quad
\textsc{int\_rate}\le25\%,\quad
\textsc{revol\_util}\le90\%,\quad
\texttt{purpose}=\texttt{small\_business}.
$
These caps align with U.S.\ underwriting rules and ensure that no training example violates regulatory or economic constraints. By contrast, unconstrained TabDiff generates only \(2\%\) valid SB loans at the 25\% rate cap, leaving AI models starved of critical high-spread examples.

We evaluate our method by comparing the performance of a CatBoost early-default classifier trained under four data regimes. The Orig-SB regime establishes a baseline using only the limited real small-business loans (\(\sim4\text{k}\) rows). The Unconstr-SB regime tests naive data augmentation by adding synthetically generated but non-compliant loans (\(\sim10\text{k}\) total rows). The Full→SB regime trains the model on the entire real loan dataset (\(\sim350\text{k}\) rows) and tests its performance on the SB segment. Finally, the CTDF-SB regime demonstrates our approach, augmenting the real SB loans with \(500\text{K}\) regulatory-compliant synthetic samples generated by CTDF.

All models are trained on pre-2018 data, validated on 2018, and
evaluated on 2019–2020Q1 SB loans only.

\begin{wrapfigure}[28]{r}{0.45\textwidth} 
  \vspace{-1.2\baselineskip}          
  \centering
  \footnotesize                       

  \setlength{\tabcolsep}{5pt}
  \renewcommand{\arraystretch}{1.12}
  \begin{tabular}{lccc}
    \toprule
    \textbf{Regime} & \textbf{AUC-PR $\uparrow$} & \textbf{F1-score $\uparrow$} & \textbf{R.@5\% FPR $\uparrow$} \\
    \midrule
    CTDF–SB   & \textbf{0.2570(18)} & \textbf{0.0305(00)} & \textbf{0.3039(13)} \\
    Uncon–SB  & 0.1820(23)          & 0.0305(00)          & 0.2093(68)          \\
    Full$\to$SB & 0.1704(12)        & 0.0734(24)          & 0.2093(21)          \\
    Orig–SB   & 0.1403(05)          & 0.0329(06)          & 0.1643(31)          \\
    \bottomrule
  \end{tabular}

  \vspace{0.55\baselineskip} 

  \includegraphics[width=\linewidth]{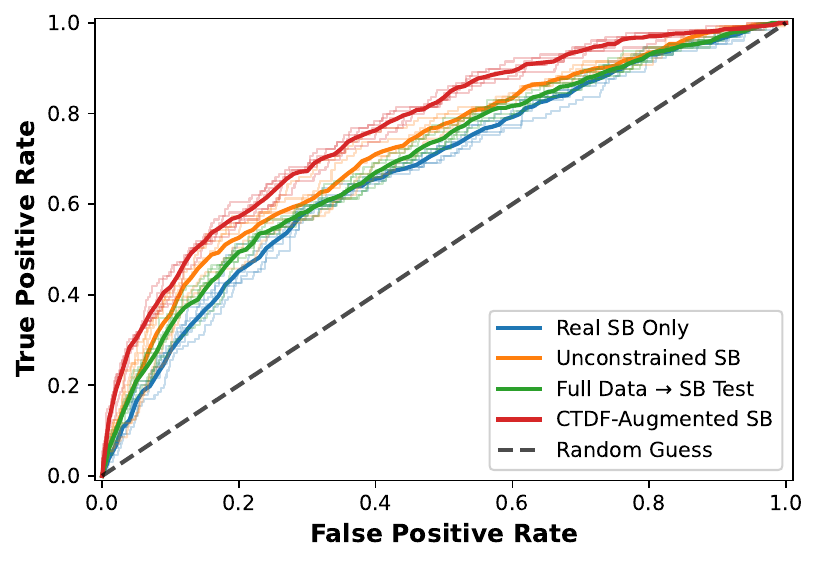}

  \caption{Top: Cross-validated performance metrics on the 2019–2020 Q1 small-business test set}
  \label{fig:sb_roc_block}
\end{wrapfigure}

In the full‐data regime, fewer than \(1\%\) of training rows
have \(\textsc{int\_rate}\ge20\%\),
so the model underfits the high-spread tail where defaults concentrate.
CTDF’s compliant augmentation supplies abundant, realistic examples
in this critical region, sharpening the AI model’s boundary and
reducing both bias and variance in predicted default risk.

\autoref{fig:sb_roc_block} demonstrates that
CTDF-SB yields the best AI performance on every metric,
e.g.\ a \emph{+0.09} uplift in AUC-PR and a 27\% relative increase in
Recall@5\% FPR versus Full→SB.  This demonstrates that
\emph{constraint-aware synthetic data} is not just compliance
insurance but \emph{AI-grade data augmentation},enabling financial
models to learn from the rare, high-risk patterns that matter most.

\paragraph{Hypothetical Scenario Planning and Extrapolation}


Financial models must often reason about scenarios absent from historical data. Our analysis of the Lending Club dataset revealed one such "data desert": no records for mortgage-holders with both a high annual income (>\$300k) and a high debt-to-income ratio (DTI > 40). To test if a generative model could fill this gap, we used CTDF for extrapolating 100k synthetic loans under these exact constraints.

\begin{wrapfigure}[20]{r}{0.47\textwidth} 
  \vspace{-9\baselineskip}          
  \centering
  \footnotesize
  \setlength{\tabcolsep}{4pt}
  \renewcommand{\arraystretch}{1.05}
  \resizebox{0.96\linewidth}{!}{%
    \begin{tabular}{@{}lrrr@{}}
      \toprule
      \textbf{Feature (Mean)} & \textbf{Real (All)} & \textbf{Real (High DTI)} & \textbf{Synth. Hyp.} \\
      \midrule
      Loan Amount (\$)     & 15,611.26 & 17,727.61 & 17,502.75 \\
      Interest Rate (\%)   & 13.00     & 15.00     & 13.00     \\
      DTI                  & 19.79     & 52.05     & 41.05     \\
      FICO Score (Low)     & 702.86    & 706.76    & 705.58    \\
      \bottomrule
    \end{tabular}%
  }

  \vspace{0.55\baselineskip} 

  \includegraphics[width=\linewidth]{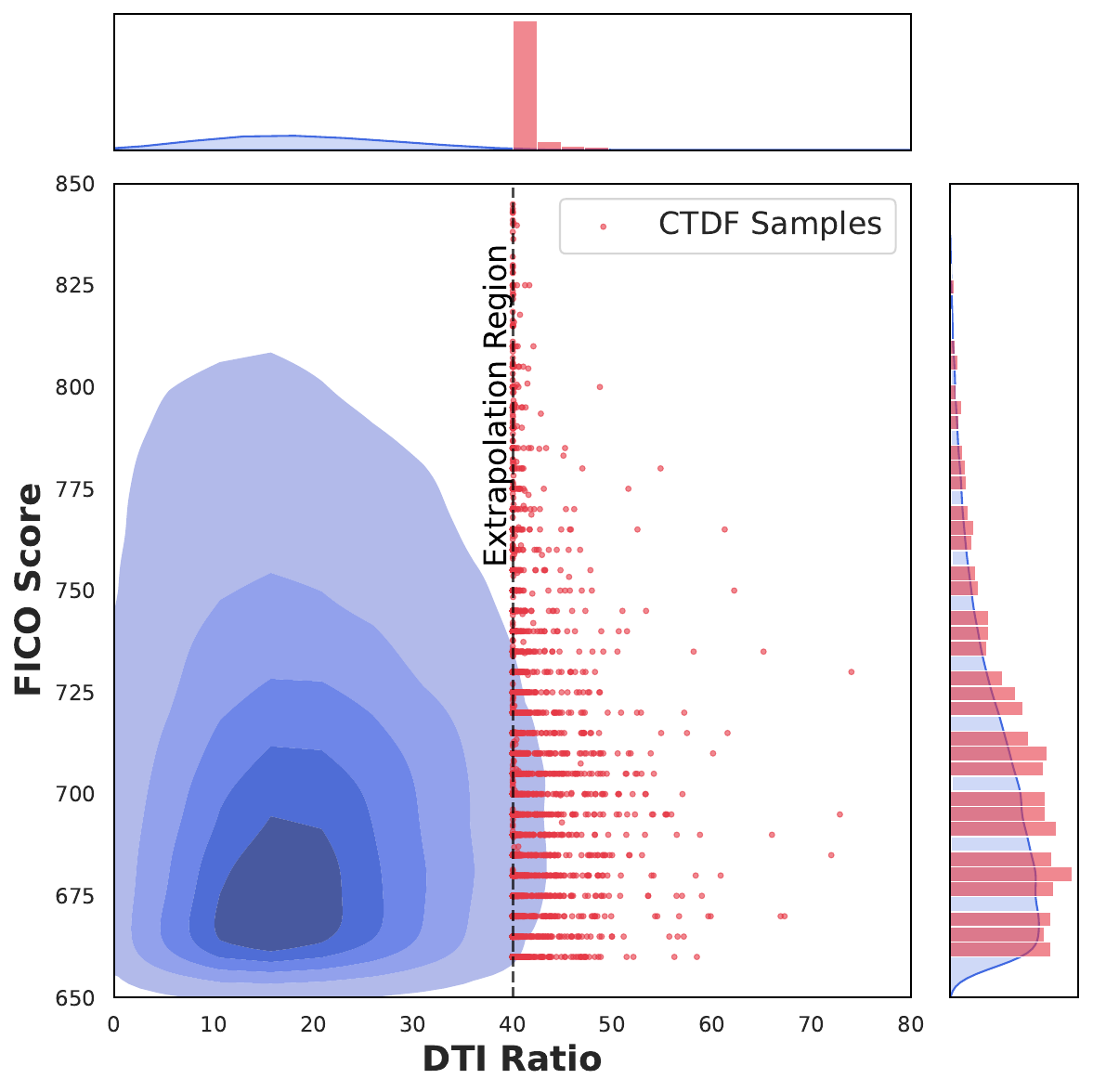}

  \caption{Top: Descriptive statistics Bottom: Realistic extrapolation of the DTI/FICO relationship; real joint density (blue) and synthetic constrained samples (red)}
  \label{fig:hypothetical_portfolio_block}
\end{wrapfigure}
 
The table results in Figure~\ref{fig:hypothetical_portfolio_block} reveal a sophisticated, non-obvious profile. While the synthetic borrowers sought large loans similar to the real high-DTI segment, the model assigned a moderate 13\% interest rate instead of a punitive one. This indicates that CTDF learned a realistic economic trade-off: the borrowers' extreme income acts as a powerful risk mitigator, offsetting their high leverage and qualifying them for more favorable terms.

By ensuring extrapolated scenarios are grounded in learned economic principles, even uncovering non-naive trends. Hence, it builds the necessary trust for financial institutions to use synthetic data for strategic decision-making.

\section{Conclusion}

We introduce Constrained Tabular Diffusion for Finance (CTDF), a training-free framework that integrates mapping‐based feasibility operations into each sampling step of a mixed-type diffusion model. By combining Euclidean metric for numerical attributes with KL-divergence for categorical variables, CTDF enforces hard affine and logical constraints with zero violations while preserving distributional fidelity and downstream utility.  Our experiments on large-scale datasets serve as illustrative case studies, showcasing CTDF’s strengths in scenario analysis, stress testing, and risk modeling. More broadly, the ability to generate compliant synthetic data opens new avenues across finance and beyond, ranging from privacy-preserving data sharing and regulatory compliance testing to algorithmic auditing, fairness evaluation, and robust AI-driven decision-making. CTDF lays the foundation for deploying constraint-aware generative modeling in diverse financial applications.

\section{Acknowledgments}

The authors gratefully acknowledge \textbf{Visa Inc.} for support through its 2025 Summer Research PhD Internship Program.

\bibliographystyle{abbrv}  
\bibliography{references} 

@misc{zhang2024tabsyn,
      title={Mixed-Type Tabular Data Synthesis with Score-based Diffusion in Latent Space}, 
      author={Hengrui Zhang and Jiani Zhang and others},
      year={2024},
      eprint={2310.09656},
      archivePrefix={arXiv},
      primaryClass={cs.LG},
      url={https://arxiv.org/abs/2310.09656}, 
}

@misc{xu2019modeling,
      title={Modeling Tabular data using Conditional GAN}, 
      author={Lei Xu and Maria Skoularidou and others},
      year={2019},
      archivePrefix={arXiv},
      primaryClass={cs.LG},
      url={https://arxiv.org/abs/1907.00503}, 
}

@misc{ishfaq2023tvae,
      title={TVAE: Triplet-Based Variational Autoencoder using Metric Learning}, 
      author={Haque Ishfaq and Assaf Hoogi and Daniel Rubin},
      year={2023},
      archivePrefix={arXiv},
      primaryClass={stat.ML},
      url={https://arxiv.org/abs/1802.04403}, 
}

@misc{kotelnikov2023tabddpm,
author = {Kotelnikov, Akim and Baranchuk, Dmitry and others},
title = {TabDDPM: modelling tabular data with diffusion models},
year = {2023},
publisher = {JMLR.org},
articleno = {725},
numpages = {16},
location = {Honolulu, Hawaii, USA},
series = {ICML'23}
}

@misc{christopher2024pdm,
      title={Constrained Synthesis with Projected Diffusion Models}, 
      author={Jacob K Christopher and Stephen Baek and Ferdinando Fioretto},
      year={2024},
      archivePrefix={arXiv},
      primaryClass={cs.LG},
      url={https://arxiv.org/abs/2402.03559}, 
}

@misc{christopher2025nsd,
      title={Neuro-Symbolic Generative Diffusion Models for Physically Grounded, Robust, and Safe Generation}, 
      author={Jacob K. Christopher and Michael Cardei and others},
      year={2025},
      archivePrefix={arXiv},
      primaryClass={cs.LG},
      url={https://arxiv.org/abs/2506.01121}, 
}

@misc{kim2023stasy,
      title={STaSy: Score-based Tabular data Synthesis}, 
      author={Jayoung Kim and Chaejeong Lee and Noseong Park},
      year={2023},
      archivePrefix={arXiv},
      primaryClass={cs.LG},
      url={https://arxiv.org/abs/2210.04018}, 
}

@misc{lee2023codi,
      title={Binning as a Pretext Task: Improving Self-Supervised Learning in Tabular Domains}, 
      author={Kyungeun Lee and Ye Seul Sim and others},
      year={2024},
      archivePrefix={arXiv},
      primaryClass={cs.LG},
      url={https://arxiv.org/abs/2405.07414}, 
}

@misc{shi2025tabdiff,
      title={TabDiff: a Mixed-type Diffusion Model for Tabular Data Generation}, 
      author={Juntong Shi and Minkai Xu and others},
      year={2025},
      archivePrefix={arXiv},
      primaryClass={cs.LG},
      url={https://arxiv.org/abs/2410.20626}, 
}

@misc{potluru2024syntheticdataapplicationsfinance,
      title={Synthetic Data Applications in Finance}, 
      author={Vamsi K. Potluru and Daniel Borrajo and others},
      year={2024},
      archivePrefix={arXiv},
      primaryClass={cs.LG},
      url={https://arxiv.org/abs/2401.00081}, 
}

@misc{lu2025machinelearningsyntheticdata,
      title={Machine Learning for Synthetic Data Generation: A Review}, 
      author={Yingzhou Lu and Lulu Chen and others},
      year={2025},
      archivePrefix={arXiv},
      primaryClass={cs.LG},
      url={https://arxiv.org/abs/2302.04062}, 
}

@misc{sohldickstein2015deepunsupervisedlearningusing,
      title={Deep Unsupervised Learning using Nonequilibrium Thermodynamics}, 
      author={Jascha Sohl-Dickstein and Eric A. Weiss and others},
      year={2015},
      archivePrefix={arXiv},
      primaryClass={cs.LG},
      url={https://arxiv.org/abs/1503.03585}, 
}

@misc{ho2020denoisingdiffusionprobabilisticmodels,
      title={Denoising Diffusion Probabilistic Models}, 
      author={Jonathan Ho and Ajay Jain and Pieter Abbeel},
      year={2020},
      archivePrefix={arXiv},
      primaryClass={cs.LG},
      url={https://arxiv.org/abs/2006.11239}, 
}

@misc{song2021scorebasedgenerativemodelingstochastic,
      title={Score-Based Generative Modeling through Stochastic Differential Equations}, 
      author={Yang Song and Jascha Sohl-Dickstein and others},
      year={2021},
      archivePrefix={arXiv},
      primaryClass={cs.LG},
      url={https://arxiv.org/abs/2011.13456}, 
}

@misc{liu2024sorareviewbackgroundtechnology,
      title={Sora: A Review on Background, Technology, Limitations, and Opportunities of Large Vision Models}, 
      author={Yixin Liu and Kai Zhang and others},
      year={2024},
      archivePrefix={arXiv},
      primaryClass={cs.CV},
      url={https://arxiv.org/abs/2402.17177}, 
}

@misc{rombach2022highresolutionimagesynthesislatent,
      title={High-Resolution Image Synthesis with Latent Diffusion Models}, 
      author={Robin Rombach and Andreas Blattmann and others},
      year={2022},
      archivePrefix={arXiv},
      primaryClass={cs.CV},
      url={https://arxiv.org/abs/2112.10752}, 
}

@misc{sahoo2024simpleeffectivemaskeddiffusion,
      title={Simple and Effective Masked Diffusion Language Models}, 
      author={Subham Sekhar Sahoo and Marianne Arriola and others},
      year={2024},
      archivePrefix={arXiv},
      primaryClass={cs.CL},
      url={https://arxiv.org/abs/2406.07524}, 
}

@misc{schiff2025simpleguidancemechanismsdiscrete,
      title={Simple Guidance Mechanisms for Discrete Diffusion Models}, 
      author={Yair Schiff and Subham Sekhar Sahoo and others},
      year={2025},
      archivePrefix={arXiv},
      primaryClass={cs.LG},
      url={https://arxiv.org/abs/2412.10193}, 
}

@misc{austin2023structureddenoisingdiffusionmodels,
      title={Structured Denoising Diffusion Models in Discrete State-Spaces}, 
      author={Jacob Austin and Daniel D. Johnson and others},
      year={2023},
      archivePrefix={arXiv},
      primaryClass={cs.LG},
      url={https://arxiv.org/abs/2107.03006}, 
}

@misc{zheng2023diffusionmodelsmissingvalue,
      title={Diffusion models for missing value imputation in tabular data}, 
      author={Shuhan Zheng and Nontawat Charoenphakdee},
      year={2023},
      archivePrefix={arXiv},
      primaryClass={cs.LG},
      url={https://arxiv.org/abs/2210.17128}, 
}

@misc{nyc_registration_law,
  author       = {{NY Mayor's Office of Special Enforcement}},
  title        = {Registration Law},
  howpublished = {\url{https://www.nyc.gov/site/specialenforcement/registration-law/registration.page}},
  year         = {2025},
  note         = {Accessed: 2025-07-12}
}

@misc{sattarov2023findiffdiffusionmodelsfinancial,
      title={FinDiff: Diffusion Models for Financial Tabular Data Generation}, 
      author={Timur Sattarov and Marco Schreyer and Damian Borth},
      year={2023},
      archivePrefix={arXiv},
      primaryClass={cs.LG},
      url={https://arxiv.org/abs/2309.01472}, 
}

@misc{seth_us_2020,
  author       = {Seth, Kritik},
  title        = {{U.S. Airbnb Open Data}},
  howpublished = {\url{https://www.kaggle.com/datasets/kritikseth/us-airbnb-open-data}},
  note         = {Accessed: 2025-07-13},
  year         = {2020},
  month        = oct,
}

@misc{ho2022classifierfreediffusionguidance,
      title={Classifier-Free Diffusion Guidance}, 
      author={Jonathan Ho and Tim Salimans},
      year={2022},
      archivePrefix={arXiv},
      primaryClass={cs.LG},
      url={https://arxiv.org/abs/2207.12598}, 
}

@misc{ethon0426_lendingclub_2007_2020q1,
  author       = {{ethon0426 (Kaggle user)}},
  title        = {Lending Club 2007–2020 Q1},
  howpublished = {\url{https://www.kaggle.com/datasets/ethon0426/lending-club-20072020q1}},
  year         = {2025},
  note         = {Accessed: 2025-07-15}
}

@book{robert1999monte,
  title={Monte Carlo statistical methods},
  author={Robert, Christian P and Casella, George and Casella, George},
  volume={2},
  year={1999},
  publisher={Springer}
}

@article{prokhorenkova2018catboost,
  title={CatBoost: unbiased boosting with categorical features},
  author={Prokhorenkova, Liudmila and Gusev, Gleb and others},
  journal={Advances in neural information processing systems},
  volume={31},
  year={2018}
}

@misc{eckerli2021generativeadversarialnetworksfinance,
      title={Generative Adversarial Networks in finance: an overview}, 
      author={Florian Eckerli and Joerg Osterrieder},
      year={2021},
      archivePrefix={arXiv},
      primaryClass={q-fin.CP},
      url={https://arxiv.org/abs/2106.06364}, 
}

@article{pu2016variational,
  title={Variational autoencoder for deep learning of images, labels and captions},
  author={Pu, Yunchen and Gan, Zhe and others},
  journal={Advances in neural information processing systems},
  volume={29},
  year={2016}
}

@misc{usdoj2025equalcredit,
  author       = {{U.S. Department of Justice, Civil Rights Division}},
  title        = {The Equal Credit Opportunity Act},
  howpublished = {\url{https://www.justice.gov/crt/equal-credit-opportunity-act-3}},
  year         = {2025},
  month        = {jan},
  day          = {2},
  note         = {Accessed: July 18, 2025}
}

@misc{dhariwal2021diffusionmodelsbeatgans,
      title={Diffusion Models Beat GANs on Image Synthesis}, 
      author={Prafulla Dhariwal and Alex Nichol},
      year={2021},
      archivePrefix={arXiv},
      primaryClass={cs.LG},
      url={https://arxiv.org/abs/2105.05233}, 
}
\end{document}